\documentclass{article}

\usepackage[final]{neurips_2023}

\usepackage[utf8]{inputenc} 
\usepackage[T1]{fontenc}    
\usepackage{hyperref}       
\usepackage{url}            
\usepackage{booktabs}       
\usepackage{amsfonts}       
\usepackage{nicefrac}       
\usepackage{microtype}      
\usepackage[dvipsnames]{xcolor}
\usepackage{graphicx}
\usepackage{wrapfig}
\usepackage{caption}
\usepackage{subcaption}

\usepackage{amsmath}
\usepackage{amssymb}

\usepackage{marginnote}

\newcommand{\Xtr}{\mathcal{X}_\mathrm{tr}}
\newcommand{\Xval}{\mathcal{X}_\mathrm{val}}

\newcommand{\subnet}[1]{g_{\leq #1}}
\newcommand{\attrxtr}{\tau_c(x_\mathrm{tr})}

\newcommand{\rulesep}{\unskip\ \vrule\ }

\title{Attributing Learned Concepts in Neural Networks\\to Training Data}

\author{%
Nicholas Konz$^{1, 2}$ \,\,
Charles Godfrey$^{4, \dagger}$ \,\,
Madelyn Shapiro$^{1}$ \,\,
Jonathan Tu$^{1}$ \\
\\
\textbf{Henry Kvinge$^{1,3}$ \,\, Davis Brown$^{1}$} \\\\
$^1$Pacific Northwest National Laboratory \,\,
$^2$Duke University\\
$^3$Department of Mathematics, University of Washington \,\,
$^4$Thompson Reuters Labs \\
}

\newcommand\blfootnote[1]{%
  \begingroup
  \renewcommand\thefootnote{}\footnote{#1}%
  \addtocounter{footnote}{-1}%
  \endgroup
}

\begin{document}

\maketitle

\begin{abstract}

By now there is substantial evidence that deep learning models learn certain human-interpretable features as part of their internal representations of data. As having the right (or wrong) concepts is critical to trustworthy machine learning systems, it is natural to ask which inputs from the model's original training set were most important for learning a concept at a given layer. To answer this, we combine data attribution methods with  methods for probing the concepts learned by a model.
Training network and probe ensembles for two concept datasets on a range of network layers, we use the recently developed \textit{TRAK} method for large-scale data attribution. 
We find some evidence for \textit{convergence}, where removing the 10,000 top attributing images for a concept and retraining the model does not change the location of the concept in the network nor the probing sparsity of the concept. This suggests that rather than being highly dependent on a few specific examples, the features that inform the development of a concept are spread in a more diffuse manner across its exemplars, implying robustness in concept formation.
  
\end{abstract}

\section{Introduction}
\vspace{-4mm}
\blfootnote{$\dagger$ Work done at Pacific Northwest National Laboratory.}
\blfootnote{Correspondence to \texttt{nicholas.konz@duke.edu} and \texttt{davis.brown@pnnl.gov}}


Given the role that concepts play in understanding and explaining human reasoning, measuring their use in neural networks is important for the goal of developing explainable and trustworthy AI. Driven by this, substantial effort has gone into developing methods that measure the presence of a concept within a neural network. Relatedly, a growing body of empirical work shows that deep neural networks learn to encode features as \textit{directions} in their intermediate hidden layers \citep{merullo2023language,wang2023concept}. A common approach to finding these directions (or \textbf{concept vectors}) is via linear probing \citep{alain2016understanding}. While probing has well-known shortcomings \citep{ravichander2020probing}, it is hard to overstate the impact that concept probing has had on deep neural network interpretability. Prominent examples include probing for syntactic concepts in `BERTology' \citep{rogers-etal-2020-primer}
and chess concepts in the AlphaZero network \citep{mcgrath2021acquisition}.

A separate thread in explainability research explores \textbf{data attribution}, which, rather than measuring the importance of a concept for a model prediction, quantifies the impact of individual \emph{training datapoints} on a given model prediction (e.g., which images in the training set were most relevant for a classifier's prediction ``zebra''?). Data attribution methods have proven to be effective for identifying brittle predictions, quantifying train-test leakage, and  tracing factual knowledge in language models back to training examples \citep{datamodels,park2023trak,grosse2023studying}. 

In this work, we explore an interplay between concept vectors and data attribution, with the goal of obtaining a better understanding of how neural networks utilize human-understandable concepts. Namely, we ask the natural question:
\vspace{-1mm}
\begin{center}
    {\em Which examples in a model's training data were important for learning hidden-layer concepts? }
\end{center}
Overall, we find that the process of learning a concept is robust to both removal of examples (no small subset of examples are critical to learning a concept-- this is consistent with the observation made in \citep{engel2023faithful} regarding attributions for the classification logit output of models on individual data points) and stable across independent model training runs. While this stability may not be surprising from a human perspective, given that concepts are, by design, supposed to be relatively unambiguous between observers, it is interesting that a similar phenomenon is seen in models.



\section{Experimental Methods}

In this section we describe the two methods that are central to our study: using a linear probe to detect a concept and then applying data attribution to concept predictions. Let $f(x)$ be an image classification neural network (the ``base model'') and assume that $f(x)$ has been pretrained on a training set $\Xtr$. Assume that $\Xval$ is the corresponding validation set. We write $f_{\leq i}$ for the composition of the first $i$ layers of $f$. Finally, for each of the concepts $c$ that we study, we assume that we have a concept training and test set $\Xtr^c$ and $\Xval^c$ where elements of these sets are labeled by whether or not the elements are examples of $c$.

\begin{figure}[h]
    \centering
    \includegraphics[width=\textwidth]{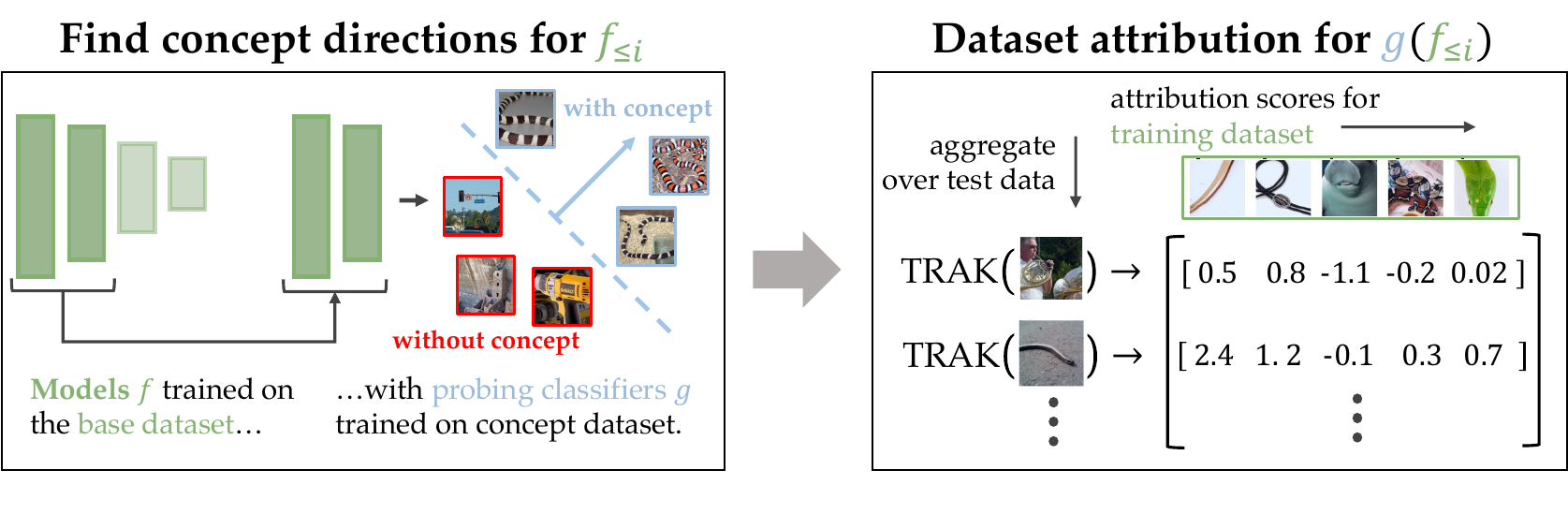}
    \caption{Schematic of our approach for concept attribution: (1) train $N$ models with different random seeds (in {\color{YellowGreen}{green}}) on the training set. (2) We choose a hidden layer $i$, append a probing classifier $g$ to its output, freeze the weights of $f_{\leq i}$, and train $(g \circ f_{\leq i})$ on the concept dataset. (3) We calculate attributions with TRAK \citep{park2023trak} for $g \circ f_{\leq i}$ on elements of the test set in terms of the original training data and aggregate across fixed layers and concepts.}
    \label{fig:attrib-summary}
\end{figure}

\subsection{Probing for Concept Learning}
\label{sec:methods:concepts}

The purpose of training a concept probe is to detect whether a specific human-interpretable concept is encoded in a hidden layer of a model. If the linear probe can effectively separate encoded exemplars of the concepts from encoded examples that are unrelated to the concept, then we take this as evidence that the model has learned the concept.
More specifically, having chosen the $i$th hidden layer of $f$ for investigation and a concept $c$ captured by concept training and test sets $\Xtr^c$ and $\Xval^c$, we follow the common approach of training an affine linear probe $g$ on the outputs of \(f_{\leq i}(\Xtr^c)\) \citep{kim2018interpretability, mcgrath2022acquisition}. 
Because $g$'s decision boundary is linear, it is effectively summarized by a normal vector which, when the probe is effective, we take to point in the ``direction'' of concept $c$ (up to a sign). This vector is called a {\emph{concept activation vector}} (CAV). 
For short-hand we will use $\subnet{i}(x) := g(f_{\leq i}(x))$ to describe the ``subnetwork + probe'' model.


\subsection{Attribution of Concept Predictions}
\label{sec:methods:attrib}

The data attribution question that we seek to answer in this paper is: ``which examples in $f$'s original training set $\Xtr$ were most important for it learning a concept $c$?'' Since we can quantify how well $f$ learned a concept at a layer $i$ by the accuracy of a trained probe $\subnet{i}(x)$ on $\Xval^c$, it is more convenient to ask ``which examples in the network's original training set $\Xtr$ were most important for the concept predictions of $\subnet{i}$ on a set of test images?''

A {\emph{data attribution method}} $\tau(x_\mathrm{val}, x_\mathrm{tr}; h) $ is a function that assigns a real-valued score to a training point $x_\mathrm{tr}\in\Xtr$ according to its importance to the prediction of a model $h$ on some test/validation point $x_\mathrm{val}$ \citep{park2023trak}. We will define the expected importance of a training point $x_\mathrm{tr}$ to concept predictions of $\subnet{i}$ as
\begin{equation}
    \label{eq:attrib}
    \attrxtr := \mathbb{E}_{x_\mathrm{val}\sim\Xval} \tau(x_\mathrm{val}, x_\mathrm{tr};\subnet{i}),
\end{equation}


as suggested by \citet{park2023trak}. For all attribution experiments we use a recently developed data attribution method, TRAK (Tracing with the Randomly-projected After Kernel) \citep{park2023trak}. For details on this method we defer to the original paper.


TRAK requires an ensemble of $M$ trained models; we use $M=20$, but found similar results for as few as $M=5$. Each model in the ensemble is a ``subnetwork+ probe'' $\subnet{i}^{(j)}$, where $1 \leq j \leq M$. $\subnet{i}^{(j)}$ is created by training the same base model $f$ on $\Xtr$ to obtain $f^{(j)}$, then training a probe for a concept $c$ on the $i^{th}$ layer of $f^{(j)}$ with the concept training set $\Xtr^c$, using $\Xval^c$ for validation. 

The first step of TRAK is to ``featurize'' (process $\Xtr$) and score (process $\Xval$) each $\subnet{i}^{(j)}$, which we run in parallel over the ensemble. After this, the attribution scores of all $M$ networks are aggregated, resulting in $\lvert \Xval  \times \Xtr \rvert $ final scores total, one for each pair $(x_\mathrm{val}, x_\mathrm{tr})$. An important note here is that TRAK requires a task/loss function and corresponding target labels to evaluate the predictions of the models $\subnet{i}^{(j)}$ --- in our case, concept prediction and binary concept labels, respectively. For consistency, unlike the concept probe training and validation sets $(\Xtr^c, \Xval^c)$ which use manually-defined concept labels, we use one of the trained $\subnet{i}^{(j)}$ to assign these labels to $\Xtr$ and $\Xval$ for attribution. We summarize our experimental design in Fig \ref{fig:attrib-summary}.

\section{Datasets, Base Models and Concepts}
\label{sec:datasets}
We use ``ImageNet10p'' as the training and validation sets $\Xtr$ and $\Xval$, respectively, to train the base models. ImageNet10p is defined by randomly sampling $10\%$ of the images of each class from the ImageNet \citep{imagenet} training and validation sets. The resulting ResNet-18 models
obtained about $45\%$ top-1 accuracy on $\Xval$ (see Appendix F for training details), from which we assume that the model learned meaningful enough visual representations for concepts to be present. We build two different concept probing datasets,\footnote{We initially experimented with using concepts from the Broden dataset \citep{bau2017network} but we found probes trained on this dataset did not generalize well to arbitrary ImageNet images.} and show example images of each concept in Appendix B. 


\paragraph{Concept 1: Snakes.}
We define the ``Snakes'' concept with the $17$ ImageNet snake classes $477$-$493$. The probe training set $\Xtr^c$ is constructed with (i) all examples of these classes in $\Xtr$ and (ii) the same number of non-snake images randomly sampled from $\Xtr$. The probe validation set $\Xval^c$ is constructed in the same manner as $\Xval$; this gives a concept training/validation split of $4,374$/$170$.

\paragraph{Concept 2: High-Low Spatial Frequencies.}
We define the ``High-Low Frequencies'' concept with images where a directional transition between a region with high spatial frequency to a region with low spatial frequency is present \citep{schubert2021high-low}.\footnote{An analogous variant of this concept was arguably also discovered in biological neurons \citep{Ding2023.03.15.532836}.} We created this concept dataset by computing the top $0.001\%$ highest-activating ImageNet images (by $L_\infty$ norm) of the `high-low frequency neurons' defined in layer {\tt mixed3a} of InceptionV1 \citep{schubert2021high-low}. 
These images are used with an equal number of random non-concept images to define the concept training and validations sets $\Xtr^c$ and $\Xval^c$, respectively, resulting in a training/validation split of $362$/$20$. We found the high selection threshold necessary to identify images where the concept is clearly present.


\section{Experiments and Results}

\subsection{Concept Attribution}

\begin{figure}[!htbp]
    \begin{subfigure}[b]{0.48\textwidth}
    \includegraphics[width=1.02\textwidth]{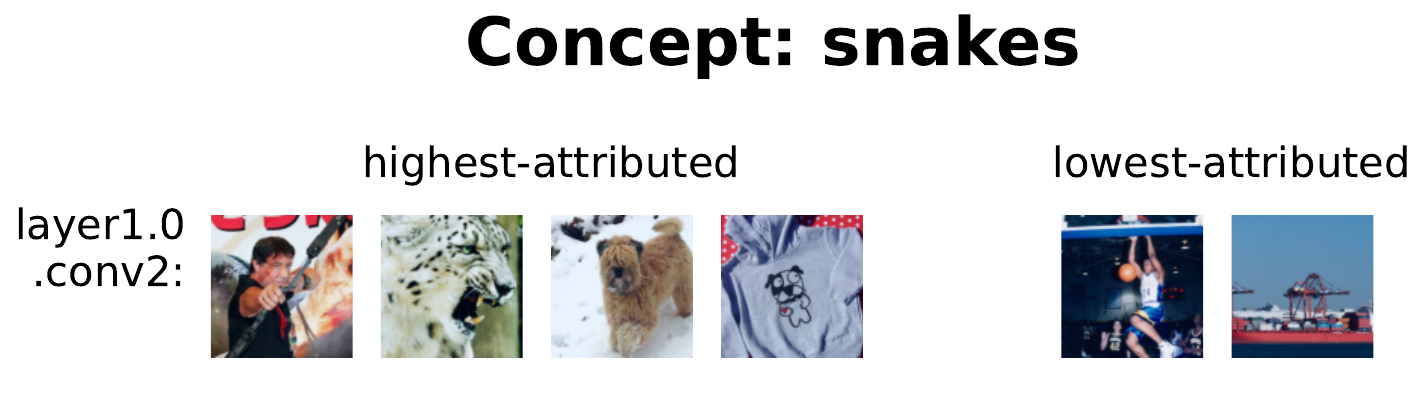}
    
    \includegraphics[width=\textwidth]{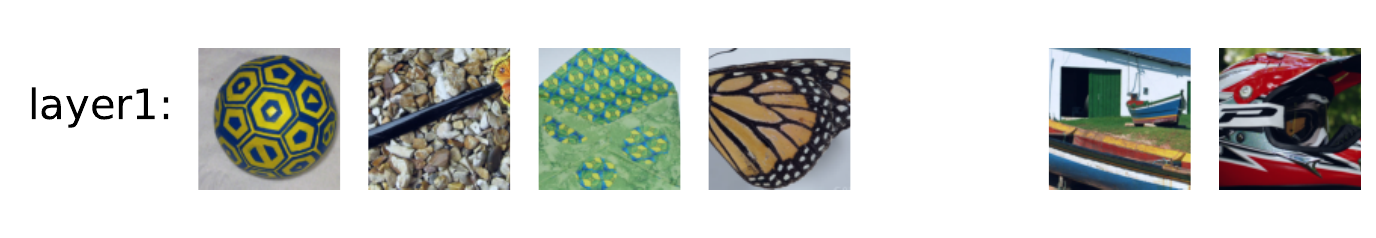}
    
    \includegraphics[width=\textwidth]{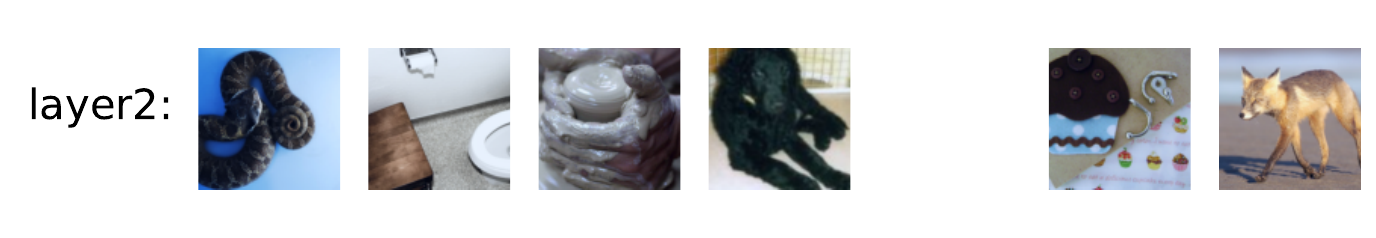}
    
    \includegraphics[width=\textwidth]{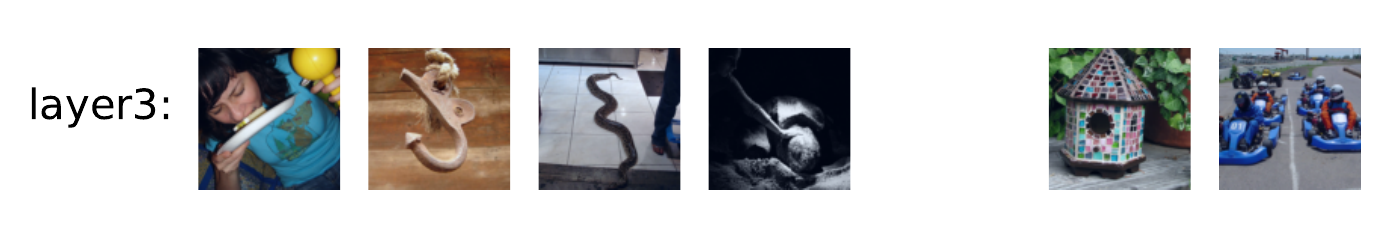}
    \end{subfigure}
    \rulesep
    \begin{subfigure}[b]{0.48\textwidth}
    \includegraphics[width=1.02\textwidth]{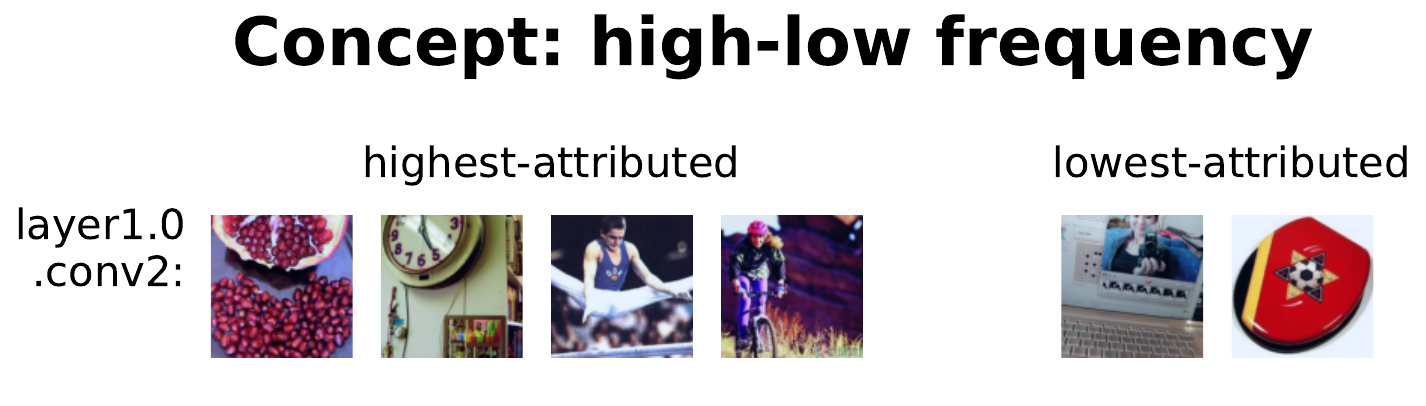}
    
    \includegraphics[width=\textwidth]{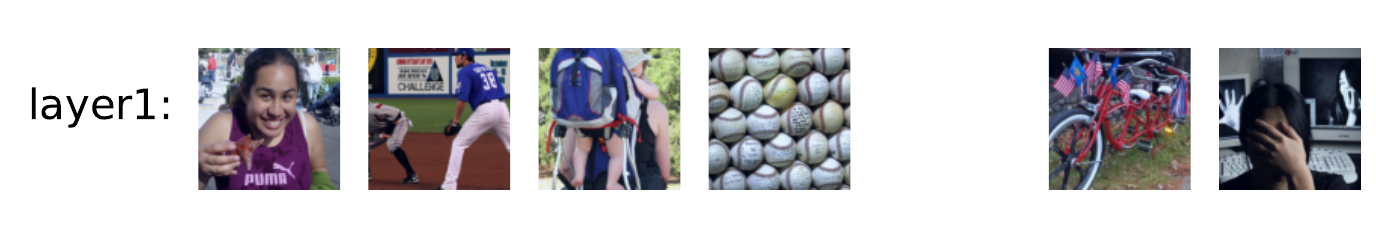}
    
    \includegraphics[width=\textwidth]{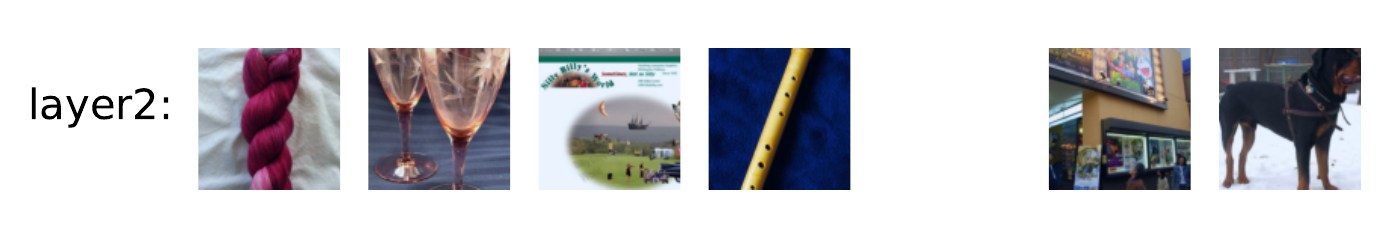}
    
    \includegraphics[width=\textwidth]{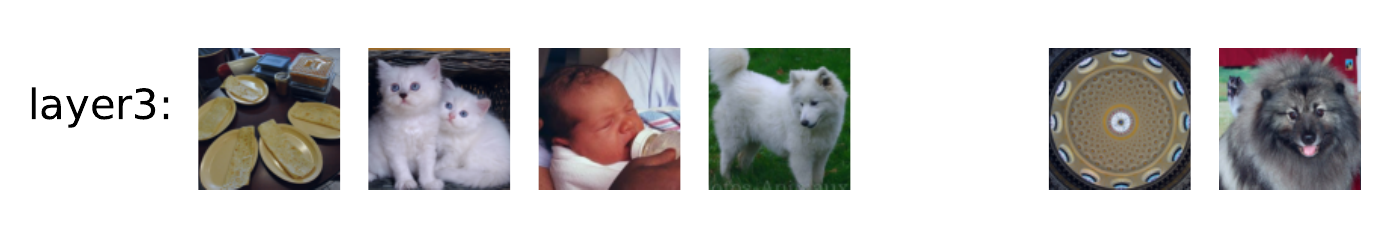}
    \end{subfigure}
    
    \caption{\textbf{Training set attributions for concept learning.} The four highest and two lowest attributed training set images (decreasing $\attrxtr$ from left to right) for concept learning at different network layers. \textbf{Left half:} snakes concept. \textbf{Right half:}  high-low frequency concept.}
    \label{fig:layerattribs}
\end{figure}
\begin{wrapfigure}[16]{r}{0.5\textwidth}
    \centering
    \includegraphics[width=\linewidth]{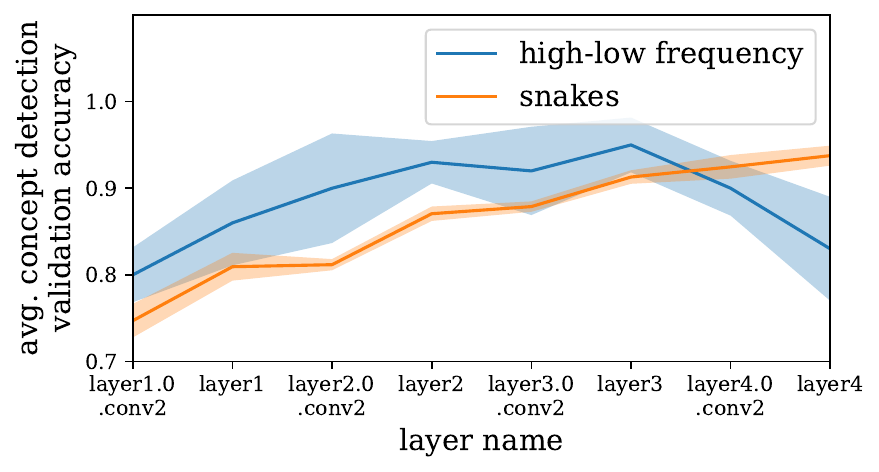}
    \caption{\textbf{Concept presence within different network layers.} Average concept detection validation accuracy of probes trained on different layers, for each concept; confidence bands are std. deviation over 5 base models.}
    \label{fig:probe_accs}
\end{wrapfigure}

In Fig. \ref{fig:layerattribs} we display the images in the 
training set $\Xtr$ which received the highest and lowest concept prediction attribution scores $\attrxtr$ (Eq. \eqref{eq:attrib}) for each concept, for various layers of the base model. In Fig. \ref{fig:probe_accs} we show how the presence of each concept varies with network layer depth, where the two concepts were most present on average in {\tt layer3}. Additional highest-attributed images are in Appendix C.1.

\paragraph{How does learned concept attribution vary between network layers?}

For the  snakes concept, full snake images appear to be important for concept learning in deeper network layers, while images that possess textures common for snakes are most important for the earlier {\tt layer1}. The concept does not appear to be present in very early layers ({\tt layer1.0.conv2}), which is reasonable given that ``snakes'' is an abstract concept (see also Fig. \ref{fig:probe_accs}). These observations are compatible with the conventional wisdom that deeper network layers learn more complex abstract features (such as objects), while earlier layers learn more basic features (such as textures).

The high-low frequency concept is fairly present throughout the network (Fig. \ref{fig:layerattribs}, right and Fig. \ref{fig:probe_accs}). Highest-attributed (and certain lowest-attributed) training set images for this concept contain transitions from high to low spatial frequency, such as pomegranate seeds over a flat background ({\tt layer1.0.conv2}, image 1), baseball threading alongside a flat casing ({\tt layer1}, image 4), interwoven threads over a smooth background, ({\tt layer2}, image 1), and fur over a smooth background ({\tt layer3}, image 2).
In comparison to the snakes concept probe which has increasing accuracy with network depth, likely due to its connection to the base models classification task, the high-low frequency concept fades after layer 3 as it is synthesized into higher-level concepts related to the label classes.

\begin{figure}
    \centering
    \begin{subfigure}[c]{0.28\textwidth}
    \includegraphics[width=\linewidth]{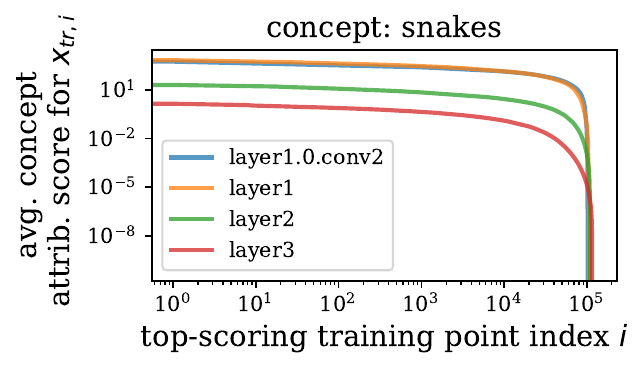}
    
    \includegraphics[width=\linewidth]{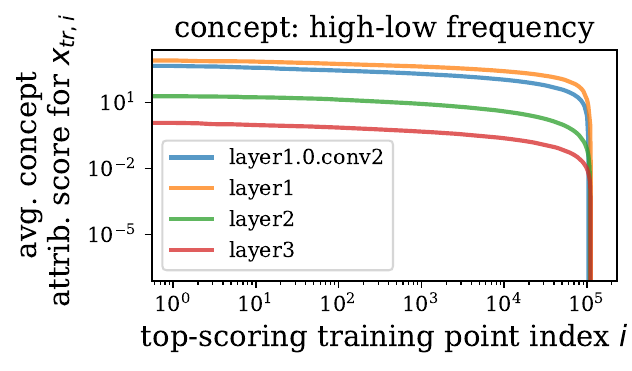}
    \end{subfigure}
    \begin{subfigure}[c]{0.48\textwidth}
    \includegraphics[width=\linewidth]{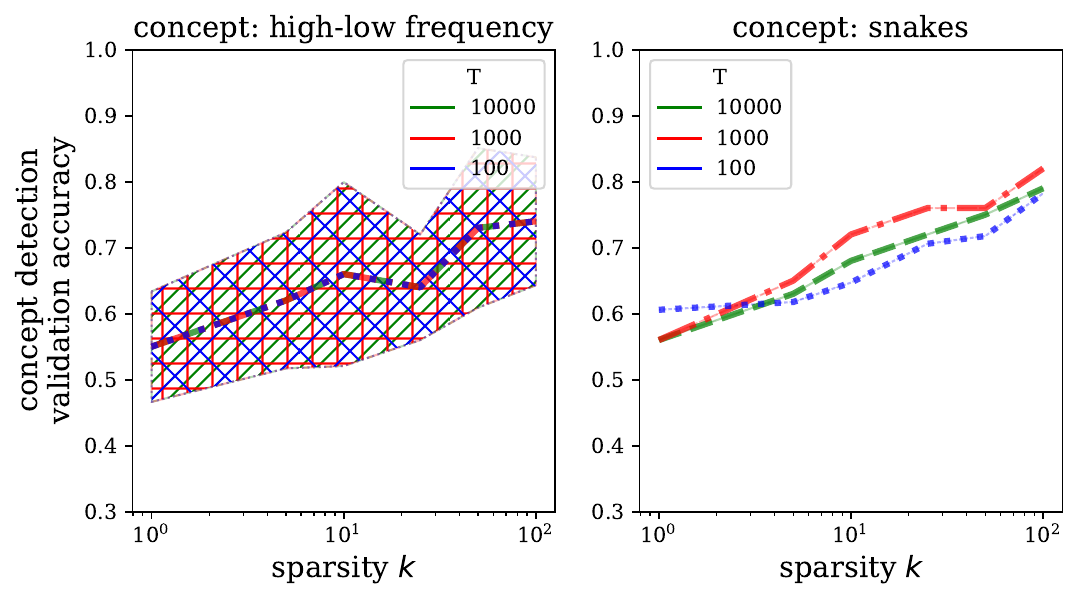}
    \end{subfigure}
    \caption{\textbf{Left:} sorted distribution of concept learning attribution scores for the training set $\Xtr$, averaged over the validation set $\Xval$, for both concepts. \textbf{Right:} Effect of re-training the base model on $\Xtr$ with the top $T$ concept-attributed training examples removed ($\Xtr^{-T}$), for sparse probe concept detection on {\tt layer3}. High-low frequency probe results averaged over 5 base models, with entirely overlapping std. deviation confidence intervals shown.}
    \label{fig:leaveout}
    \vspace{-1em}
\end{figure}

\paragraph{Are the concepts that a model learns the result of a few select exemplars?}
We analyze the importance of images in the base model's training set $\Xtr$ for concept learning by (1) removing the $T$ highest-attributed images from $\Xtr$ to obtain $\Xtr^{-T}$ ($T\in\{100, 1,000, 10,000\}$), (2) re-training the $M$ base models on $\Xtr^{-T}$, and (3) training concept probes on each of them for a given layer.\footnote{
Concept probe training and validations sets $\Xtr^c$, $\Xval^c$ are unchanged from their original definition (Sec. \ref{sec:datasets}).} If the probe concept detection validation accuracy changes after the training set is pruned of the most important examples of a concept, then we conclude that these examples were primarily responsible for the model learning the concept. If this does not happen, it suggests that a model learns a concept in a more flexible way, from a broad range of examples. For this experiment we measure concepts in the layer where both were most present on average, {\tt layer3}. Our results are shown in the middle and right plots in Figure \ref{fig:leaveout} (where these first experiments correspond to sparsity equal to $10^2$), concept validation accuracy did not change on models trained on $\Xtr^{-T}$ for varying $T$ compared to the baseline of those on $\Xtr$, for either concept (Figure \ref{fig:probe_accs}).\footnote{High-low frequency probes at different $T$ for the same base model obtained the same performance due to the small size of $\Xval^c$.} This provides further evidence that the learning of a concept is diffuse among exemplars and does not depend on a few special examples. We note that this result may not be surprising given that the attribution scores are mostly similar across examples from $\Xtr^c$, Figure \ref{fig:leaveout} left, (e.g., if all examples have the same importance for learning a concept, removing a fraction of them will not have a large effect). In particular, the high-low frequencies concept could be learnable from the majority of images in the training set, especially if it suffices for the probes simply to learn to be boundary detectors; we investigate this with ``relative probing'' to discriminate between generic object boundaries and the high-low frequency concept in Appendix D.

\begin{wrapfigure}[10]{r}{0.45\textwidth}
    \vspace{-1.5em}
    \centering
    \includegraphics[width=\linewidth]{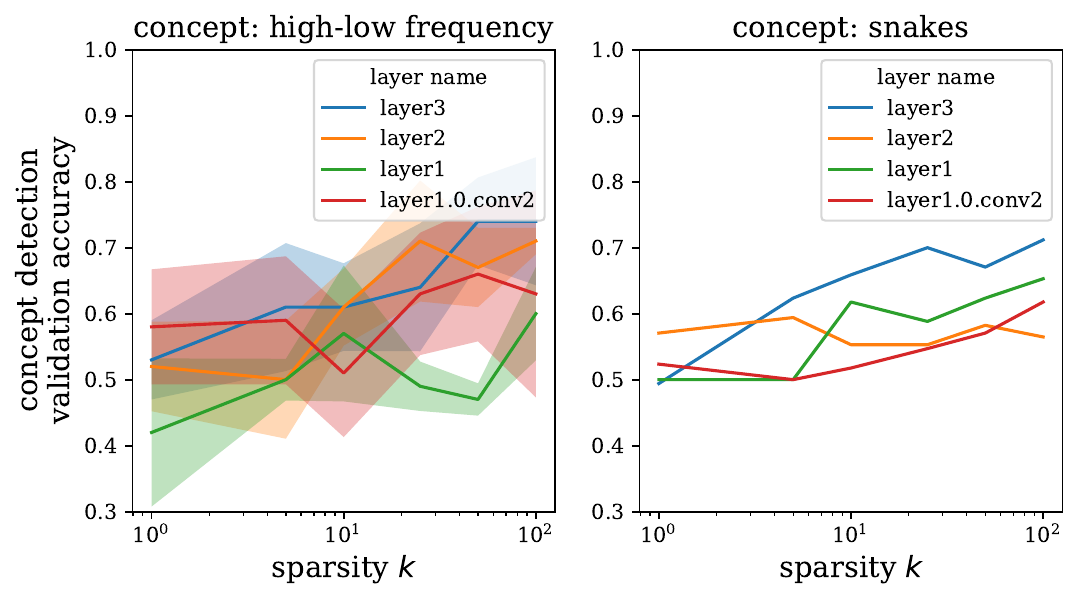}
    \vspace{-1.8em}
    \caption{Concept detection validation accuracy vs. concept probe sparsity $k$.}
    \label{fig:sparsity}
\end{wrapfigure}

Finally, given evidence that semantically meaningful representations tend to be sparse in the neuron basis \citep{gurnee2023finding}, we also trained sparse probes, thinking that in this regime removing a fraction of exemplars might have a larger effect. In the middle and right plots in Figure \ref{fig:leaveout} and Figure \ref{fig:sparsity} the concepts also remained robust in this setting.


\paragraph{How similar are different probes trained for the same concept/layer?} 


We show how probe accuracy changes with respect to network layer depth in Fig. \ref{fig:probe_accs}. For a fixed layer, different probes typically converge to similar performance.\footnote{In Appendix E we discuss why we do not compare probes via CAV similarity.}

In the case of the high-low frequency concept, we see that probe accuracy is highest at intermediate layers, and comparatively low at the earliest and latest layers. This is consistent with the original work of \cite{schubert2021high-low}, which discovered ``high-low frequency detector neurons'' in intermediate layers of InceptionV3 networks (but not in the earliest or latest layers). This stands in contrast to the snakes concept, where probe accuracy increases monotonically with network depth. One possible explanation for this observation is that the snakes concept dataset was obtained from a subset of ImageNet classes. As such, the base models have been trained to correctly classify positively-labelled concept images using their output logits.
Here our observations are consistent with \cite{alain2016understanding}, which trained multi-class linear probes for ImageNet classification and found monotonically increasing accuracy with depth.

\subsection{Sparse Concept Probing and Attribution}

\paragraph{How does forcing probes to be sparse affect concept detection?}
Forcing a probe to be sparse (at most $k$ of the CAV elements are non-zero) allows for even more interpretable concept directions \citep{gurnee2023finding}. To evaluate this for our concepts, we trained probes with a range of sparsities on different layers of $f$, using an approach similar to iterative hard thresholding \citep{jin2016training}.
After the first half of the training epochs, we set all but the $k$ parameters of highest absolute value to zero, freeze all of the zeroed parameters from updating, then continue training.

In Fig \ref{fig:sparsity} we show how the concept detection validation accuracy changes with probe sparsity $k$ at multiple layers, for both concepts. Reasonably, probe accuracy typically increases with $k$, and we see a similar relative accuracy ranking of different layers as in the non-sparse case (Fig. \ref{fig:probe_accs}). We see that the concepts are both somewhat learnable with sparse probes, but not nearly to the degree of the non-sparse probes (Fig. \ref{fig:probe_accs}).

\begin{figure}[!htbp]
    \centering
    \includegraphics[width=0.95\textwidth]{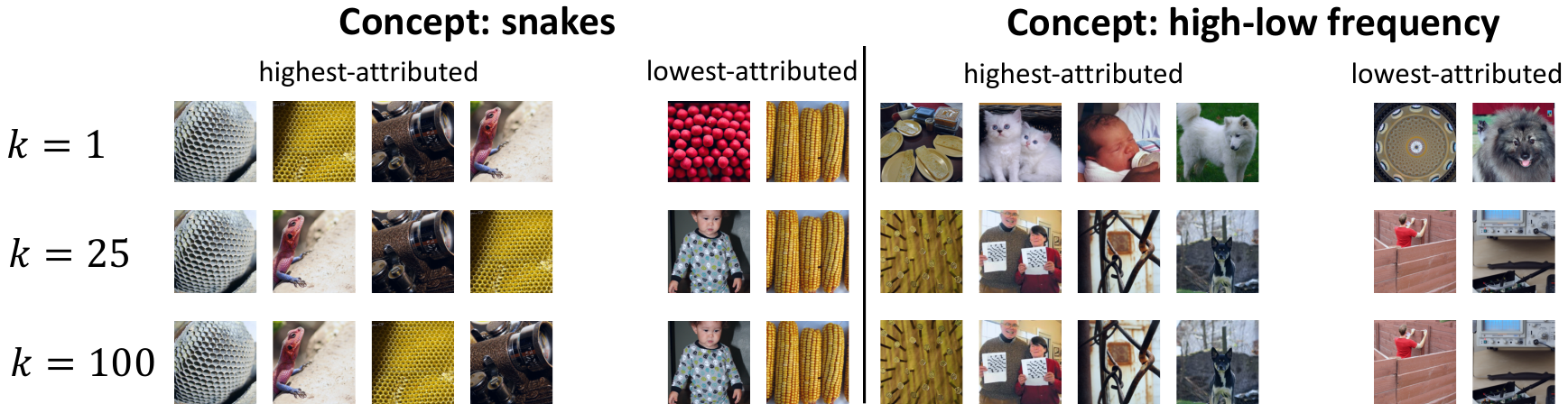}
    
    \caption{\textbf{Training set attributions for sparse probe concept learning.} The four highest and two lowest attributed training set images for concept learning at {\tt layer3} for probes of different sparsity $k$. \textbf{Left half:} snakes concept. \textbf{Right half:}  high-low frequency concept.}
    \label{fig:sparseattribs}
\end{figure}

Fig. \ref{fig:sparseattribs} displays the highest- and lowest-attributed images for probes of varying sparsity $k$, for each concept at {\tt layer3} (compare with the attributions of non-sparse probes in Fig. \ref{fig:attrib-summary}). For the snakes concept, the attributions are different than those for the non-sparse probe, and yet very similar among the sparse probes. Interestingly, we see that almost all of the highest- or lowest-attributed images possess a ``honeycomb''-like texture which appears similar to snake scales. For the high-low frequencies concept, we see that the $k=1$ sparse probe obtained the same attributions as the non-sparse probe, yet the probes with more non-zero entries both obtained the same distinct attributions, which also appear to have examples of the concept. 

\section{Limitations}
In order to experimentally vary factors including model layer, concept and concept training data, we were forced to restrict other variables. We only experiment with ResNet18 image classifiers trained on ImageNet10p and two concept datasets (snakes and high-low frequency) --- adding additional base models and concepts would increase the breadth of this study. Another interesting future direction would be conducting analogous experiments on a different modality (e.g., natural language) or task. Finally, we only use TRAK for data attribution, and it would be interesting to know the extent to which our experimental results are particular to TRAK.

\section*{Conclusion}

In this paper we explored the importance of individual datapoints in concept learning. We found evidence of ``convergence'' in several senses, including stability under removal of exemplars and across independent training runs. Although more extensive experiments are needed (with better aggregation methods), our results suggest a robustness to the way that concepts are learned and stored in a deep learning model.

\section*{Acknowledgments}
This research was supported by the Mathematics for Artificial Reasoning in Science (MARS) initiative at Pacific Northwest National Laboratory.
It was conducted under the Laboratory Directed Research and Development (LDRD) Program at at Pacific Northwest National Laboratory (PNNL), a multiprogram
National Laboratory operated by Battelle Memorial Institute for the U.S. Department of Energy under Contract
DE-AC05-76RL01830.

The authors would also like to thank Andrew Engel for useful conversations and feedback related to the paper.


\bibliographystyle{plainnat}
\bibliography{refs}

\clearpage
\appendix

\section{Related Work}

\textbf{Concept vectors:} Using linear probes to study the hidden features of neural networks dates back at least as far as \citep{alain2016understanding}, although in that work, the probes were performing the same task used to train the underlying neural network. The framing of concept vectors originates in \citep{kim2018interpretability}, which demonstrated their usefulness for making neural networks more interpretable.
There has been a large amount of follow-up work concerning concept vectors, neuron interpretability, and linear representations. 



\textbf{Data attribution:} Our primary references are the \emph{datamodels} framework \citep{datamodels} and its more computationally-tractable approximation TRAK \citep{park2023trak}. For a more thorough discussion of related work on data attribution (which is a classical topic in statistical learning), we refer to these two papers.

\section{Example Concept Images}
\label{app:egconcepts}

\begin{figure}[!htbp]
    \centering
    \includegraphics[width=0.95\linewidth]{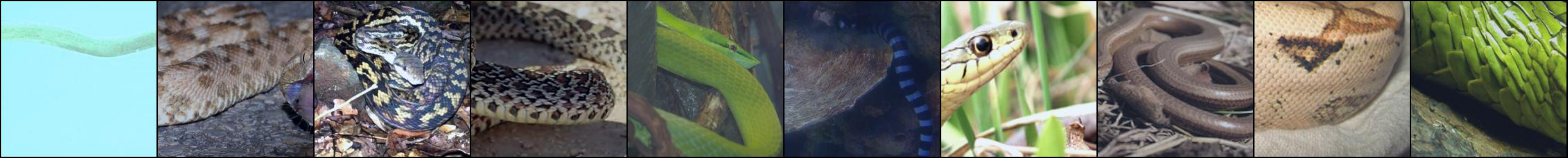}
    \includegraphics[width=0.95\linewidth]{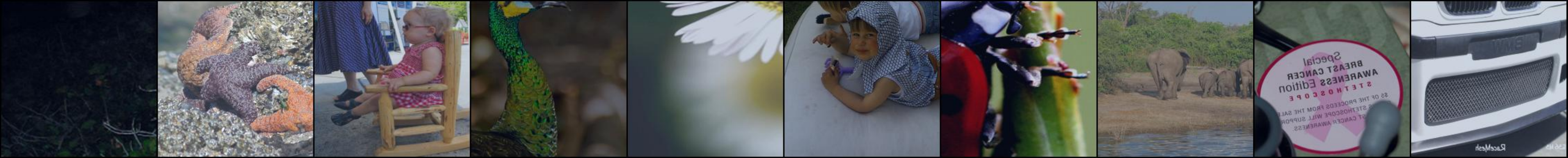}
    \caption{Example images from our concept datasets. \textbf{Top:} Snakes; \textbf{Bottom:} High-Low frequencies.}
    \label{fig:app:eg_concepts}
\end{figure}

\section{Additional Results}

\subsection{Highest-Attributed Training Set Images for Concept Learning}
\label{app:moreattrib}

\begin{figure}[!htbp]
    \centering
    \includegraphics[width=0.95\linewidth]{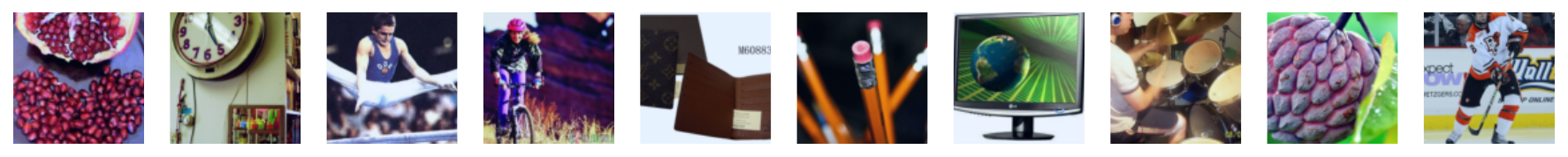}
    \includegraphics[width=0.95\linewidth]{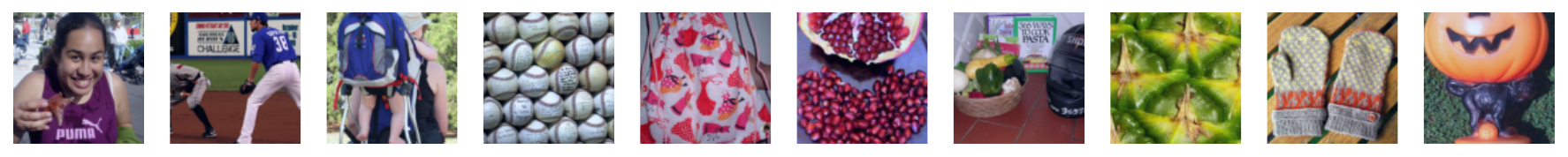}
    \includegraphics[width=0.95\linewidth]{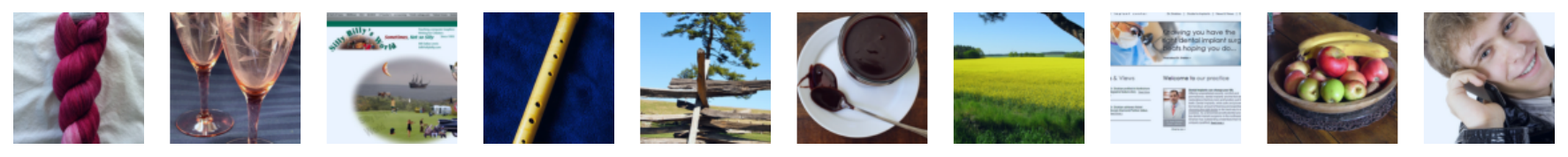}
    \includegraphics[width=0.95\linewidth]{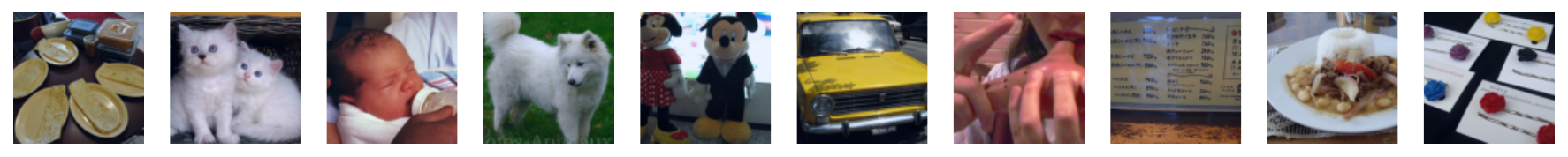}
    \caption{For the high-low frequency concept, the ten highest attributed training set images (decreasing $\attrxtr$ from left to right) for concept learning at different network layers. \textbf{From top to bottom:} {\tt layer1.0.conv2}, {\tt layer1}, {\tt layer2}, {\tt layer3}.}
    \label{fig:app:moreattrib_hilow}
\end{figure}

\begin{figure}[!htbp]
    \centering
    \includegraphics[width=0.95\linewidth]{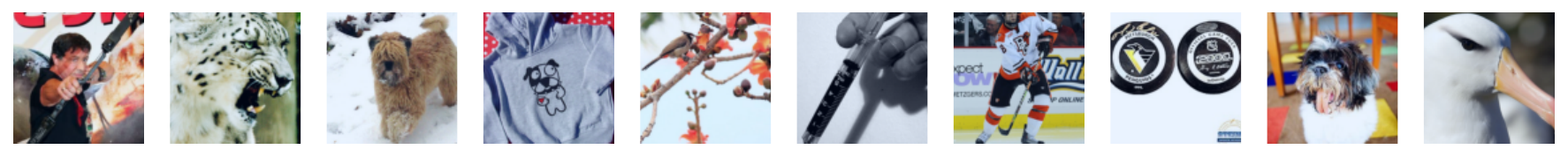}
    \includegraphics[width=0.95\linewidth]{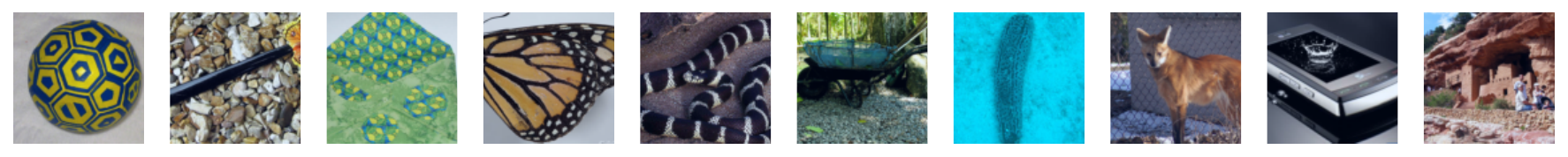}
    \includegraphics[width=0.95\linewidth]{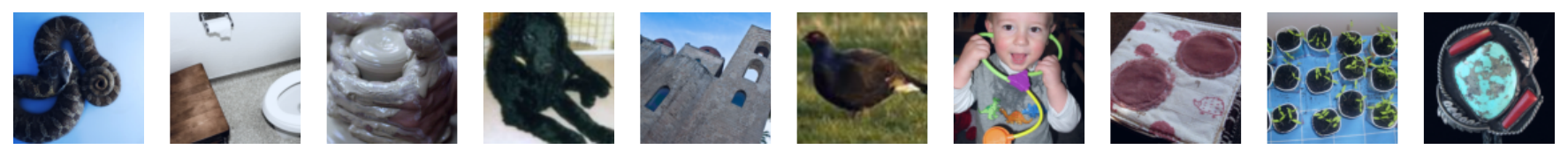}
    \includegraphics[width=0.95\linewidth]{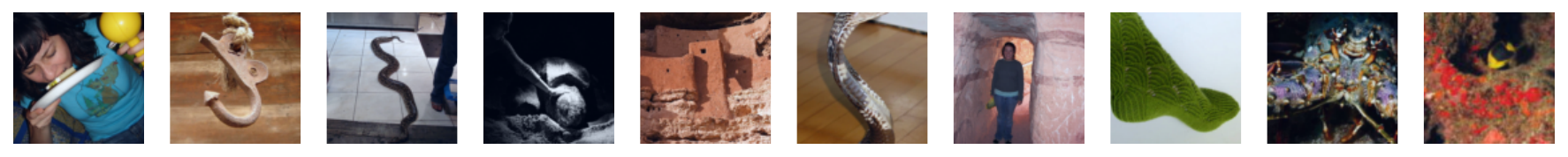}
    \caption{For the snakes concept, the ten highest attributed training set images (decreasing $\attrxtr$ from left to right) for concept learning at different network layers. \textbf{From top to bottom:} {\tt layer1.0.conv2}, {\tt layer1}, {\tt layer2}, {\tt layer3}.}
    \label{fig:app:moreattrib_snakes}
\end{figure}

\section{Relative Concept Probing: High-Low Frequencies vs. Boundaries}

In this section, we explore probing for the ``High-Low Frequencies'' concept further, to attempt to judge if the probes can actually detect the concept, or are relying on the simpler task of boundary detection. To do so, we train probes to differentiate between the high-low frequencies concept and a concept of boundaries, as follows. 

We create the boundaries concept with a similar procedure to the creation of the high-low concept (Sec. \ref{sec:datasets}): we take the top $0.001\%$ highest-activating ImageNet training images in for the 57 neurons identified as boundary or curve detectors in {\tt mixed4b} of InceptionV1 \citep{schubert2021high-low}, to define the boundary images for concept training, and we do the same from the ImageNet validation set (using the activation percentile created from the training set) to define the boundary images used for concept validation. The final concept training and validation sets $\Xtr^c$ and $\Xval^c$, respecitvely, are then created by combining these boundary concept images with the respective high-low frequency concept training or validation images, and random-sampling from the concept with more examples to make the final $\Xtr^c$ and $\Xval^c$ equally class-balanced. This results in a concept train/validation split of $362$/$20$.

\begin{figure}[htbp!]
    \centering
    \includegraphics[width=0.5\linewidth]{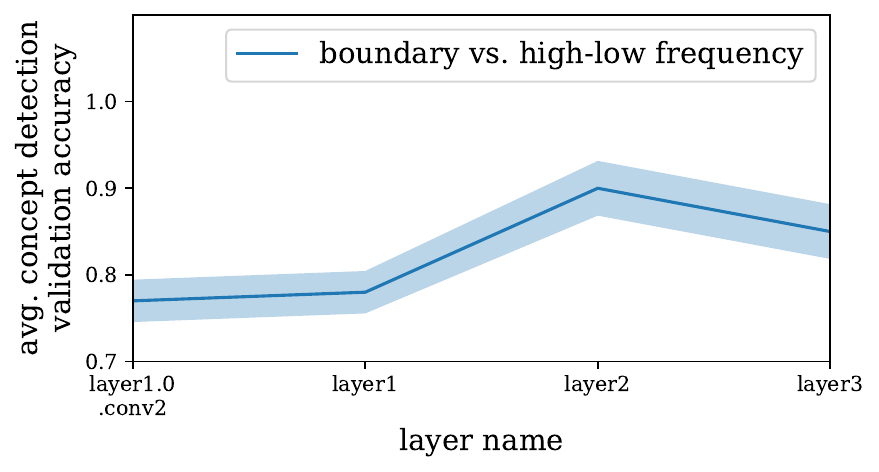}
    \caption{Average concept detection validation accuracy of probes trained on different layers, for boundary vs. high-low frequency concept classication; confidence bands are std. deviation over 5 base models.}
    \label{fig:app:bdryhilow_probeaccs}
\end{figure}

\begin{figure}[htbp!]
    \centering
    \includegraphics[width=0.78\linewidth]{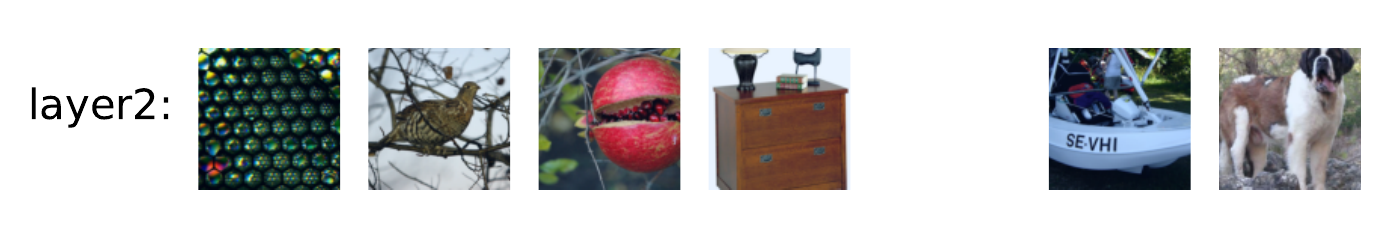}
    \caption{Four highest- (\textbf{left}) and two lowest-attributed (\textbf{right}) images in the base model training set $\Xtr$ for relative probing of boundary vs. high-low frequency concepts in {\tt layer 2} of the base model.}
    \label{fig:app:bdryhilow_attrib}
\end{figure}

After training probes to differentiate between these two concepts, we see that across the network depth, they are able to succeed ($\sim75$--$90\%$ accuracy on $\Xval^c$), especially in {\tt layer2} (Fig. \ref{fig:app:bdryhilow_probeaccs}). This leads us to believe that our high-low frequency probes can indeed tell the difference between that more complex concept than the ``proxy'' of simple boundaries. The base model training set attributions for this relative probing are also interesting (Fig. \ref{fig:app:bdryhilow_attrib}). Top-attributed images have clear examples of high-low frequency transitions (top three images) or boundaries (fourth-highest image); intuitively the type of images from which the differentiation between these two concepts could be learned.

\section{Additional Discussion}
\label{app:disc}
\paragraph{Why we compare probes with concept detection accuracy instead of CAV similarity.}

Conceivably, we could compare the concept activation vectors (CAVs) of the $M$ different probes+base models trained for the same concept and layer in order to see how much the learned concept direction varies, e.g. with the cosine similarity of two probe CAVs. However, as each probe is trained on a different base model, the activations $f_{\leq i}(x)$ in the given layer for each model may possess different bases, such that comparing the probe CAVs directly may not be informative. If we assume the activation bases to only vary by some rotation $U$, we can instead simply compare the probe predictions, as follows. Consider two trained base networks $f^{(1)}$ and $f^{(2)}$ with CAVs $a^{(1)}$ and $a^{(2)}$, respectively, such that $f^{(1)}_{\leq i}(x) = U f^{(2)}_{\leq i}(x)$ and $a^{(1)} = U a^{(2)}$. Then $(a^{(1)})^Tf^{(1)}_{\leq i}(x) = (U a^{(2)})^T U f^{(2)}_{\leq i}(x) = (a^{(2)})^T U^T U f^{(2)}_{\leq i}(x) = (a^{(2)})^T f^{(2)}_{\leq i}(x)$, since $U^T U=I$. Therefore, the probe logits $(a^{(j)})^T f^{(j)}_{\leq i}(x)$ (and predictions, ignoring bias terms) can all be compared directly. 

\section{Base Model and Concept Probe Training Details}

For training our base ResNet18s, we use the FFCV library \citep{leclerc2023ffcv} --- our training code is a fork of their \href{https://github.com/libffcv/ffcv-imagenet/}{ImageNet demo} with the following changes: 

\begin{itemize}
    \item We drop in our ImageNet10p (and for the ``drop top-$T$ highest attribution images'' experiments, the appropriate subsets of ImageNet10p) datasets.
    \item We create a variant of their \texttt{rn18\_88\_epochs.yaml} 88-epoch ResNet18 training configuration with a lower base learning rate (0.125 as opposed to 0.5), lower batch size (256 as opposed to 1024) and greater total number of epochs (176 as opposed to 88).
\end{itemize}

The batch size was decreased to facilitate training on NVIDIA V100 GPUs (as opposed to A100s, which the original training config targeted) and the learning rate was decreased proportionally. The total number of epochs was doubled to ensure we trained to convergence, i.e. to be more confident that the low accuracy (\(\approx 45\%\)) was due to training on only 10\% of ImageNet, and not a result of under-optimization. For each variant of ImageNet10p, we trained 20 ResNet18s from different random seeds in parallel on a GPU cluster.

We train concept probes with a binary cross-entropy loss and Adam \citep{adam}, using a learning rate of $5 \times 10^{-5}$, batch size of $64$, and weight decay of $10^{-5}$, for 20 and 50 epochs for our two concepts (snakes and high-low frequencies), respectively. We use standard image augmentations during probe training. 

\end{document}